%% file: Main.tex
\documentclass[10pt,journal]{IEEEtran}
\ifCLASSOPTIONcompsoc
  \usepackage[nocompress]{cite}
\else
  \usepackage{cite}
\fi
\input{Config.tex}

\begin{document}
\title{M2RU: \underline{M}emristive \underline{M}inion \underline{R}ecurrent \underline{U}nit for On-Chip Continual Learning at the Edge}


\author{Abdullah M. Zyarah and
        Dhireesha Kudithipudi,~\IEEEmembership{Senior member,~IEEE ~\vspace{-5mm}}%
\IEEEcompsocitemizethanks{
\IEEEcompsocthanksitem Abdullah M. Zyarah is with the NUAI Lab, Department of Electrical and Computer Engineering, University of Texas at San Antonio and Department of Electrical Engineering, University of Baghdad (E-mail: abdullah.zyarah@uob.edu.iq).
\IEEEcompsocthanksitem Dhireesha Kudithipudi is with the NUAI Lab, Department of Electrical and Computer Engineering, University of Texas at San Antonio, TX 78249 USA (E-mail: dk@utsa.edu).
}}

\markboth{2026}%
{Shell \MakeLowercase{\textit{et al.}}: Bare Demo of IEEEtran.cls for Computer Society Journals}

\IEEEtitleabstractindextext{%
\begin{abstract}

Continual learning on edge platforms remains challenging because recurrent networks depend on energy-intensive training procedures and frequent data movement that are impractical for embedded deployments. 
This work introduces M2RU, a mixed-signal architecture that implements the minion recurrent unit for efficient temporal processing with on-chip continual learning. The architecture integrates weighted-bit streaming, which enables multi-bit digital inputs to be processed in crossbars without high-resolution conversion, and an experience replay mechanism that stabilizes learning under domain shifts. M2RU achieves 15 GOPS at 48.62 mW, corresponding to 312 GOPS per watt, and maintains accuracy within 5 percent of software baselines on sequential MNIST and CIFAR-10 tasks. Compared with a CMOS digital design, the accelerator provides 29$\times$ improvement in energy efficiency. Device-aware analysis shows an expected operational lifetime of 12.2 years under continual learning workloads. These results establish M2RU as a scalable and energy-efficient platform for real-time adaptation in edge-level temporal intelligence.
\end{abstract}


\begin{IEEEkeywords}
M2RU, Minion Recurrent Unit (MiRU), Neuromorphic hardware, On-chip Continual learning, DFA for RNN~\vspace{-2mm}
\end{IEEEkeywords}}

\maketitle
\IEEEdisplaynontitleabstractindextext
\IEEEpeerreviewmaketitle

{\section{Introduction}\label{sec:introduction}}
\IEEEPARstart{C}{ontinual} learning is a fundamental requirement for  systems that operate in dynamic and real-world environments. Unlike ML training paradigms that assume static data distributions and repeated offline retraining, continual learning models should adapt as new information becomes available while preserving prior knowledge~\cite{parisi2019continual}. This ability seems to be essential for several edge applications, such as sensing, autonomous driving, that handle streaming data under tight energy and latency constraints~\cite{yin2023continual}. However, several of the existing models still rely on centralized training pipelines that require substantial computation and data movement, which limits their application to edge deployments. 

On the other hand, recurrent neural networks (RNNs) naturally model temporal and sequential structure, yet their integration into continual learning frameworks remains limited. Previous studies have explored only a narrow set of hardware architectures, including continual variants of echo state networks~\cite{cossu2021continual, yang2025continual}, LSTMs~\cite{kinoyama2023continual}, and GRUs~\cite{zyarah2025minion}. However, these efforts focus almost exclusively on algorithmic techniques and provide little support for continual adaptation at the hardware level. In practice, most systems still follow a train-then-deploy paradigm in which training is performed on centralized servers, while inference is executed on edge devices with severe constraints on compute, memory, and energy. This rigid separation prevents models from adapting to evolving data distributions and leads to deterioration of performance over time, raising reliability concerns in dynamic real world settings~\cite{ravaglia2021tinyml}.

Recent efforts have introduced memristor-based neuromorphic implementations to alleviate these issues by integrating non-volatile memory and computation. Early research focused primarily on memristive LSTM architectures. For example, Dou et al. proposed a memristor-based LSTM network for text classification from the IMDB dataset. The network used a 1T1R crossbar architecture to reduce slip path currents and was trained ex-situ~\cite{dou2023memristor}. Building on this direction, subsequent work explored modifications to gate structures to reduce complexity. Xiangrong et al. proposed a hybrid memristor-based GRU, which fuses the update and reset gates as originally suggested by~\cite{zhou2016minimal} to enhance computational speed and reduce storage requirements~\cite{pu2025memristor}. The design is validated with the MNIST and IMDB datasets. While these approaches demonstrate the feasibility of mapping gated RNNs to memristive architectures, they remain dependent  on ex-situ training and are limited to static tasks. To overcome this limitation, Liu et al. introduced a memristor-based implementation of LSTM capable of performing in-situ using backpropagation, validated in sentiment and digit classification using MNIST and IMDB datasets~\cite{liu2020memristor}. Zhang et al. introduced a memristor-based GRU network to recognize hand-written characters from the CASIA dataset. The authors used separated memristors within each GRU cell to emulate synaptic weights rather than integrating them into a crossbar structure, leading to an inevitable increase in the utilized resources. More recently, Zyarah et al. introduced a memristive reservoir system for time-series forecasting and sequence learning, which offers in-situ training and robustness to device failure when evaluated against univariate benchmarks~\cite{zyarah2024time}. 

Despite these advances, there are no existing memristor-based gated RNN that supports on-chip continual learning, with mechanisms that safeguard previously acquired knowledge while processing new temporal data. To address this gap, this work presents M2RU, a memristor-based mixed-signal implementation of the Minion Recurrent Unit (MiRU), designed for efficient and adaptive temporal learning directly on-chip. The contributions of this work are as follows:

\begin{itemize}
  \item   \textbf{A mixed signal architecture} of the Minion Recurrent Unit, providing a compact alternative to GRU based temporal processing with reduced gate complexity and lower storage requirements, with $\approx$ 5 percent accuracy degradation, and an estimated operational lifetime of 12.2 years.

\item \textbf{A weighted bit streaming method} for crossbar computation, which assigns bit significance using memristor based ratio scaling and eliminates the need for high resolution ADCs/DACs during multi-bit input presentation.

\item \textbf{An on-chip continual learning framework} based on direct feedback alignment, enabling efficient weight updates without backward locking or storage of intermediate states, and supporting temporal adaptation under domain incremental scenarios.

\item \textbf{A hardware integrated experience replay mechanism}, including a reservoir sampler and stochastic quantizer optimized for nonstationary data streams with minimal memory footprint.

\end{itemize}



The rest of the paper is organized as follows: Section II discusses the basics of continual learning, MiRU network architecture, and training. Section III presents the system design and implementation of the MiRU RNN. The design methodology is introduced in Section IV. Sections V and VI, respectively, discuss the results and conclude the paper.

\section{Background}
\subsection{Continual Learning}
Continual learning refers to the ability of the system to continuously acquire new knowledge from non-stationary data without experiencing catastrophic forgetting. Catastrophic forgetting usually occurs due to the lack of balance between the stability and plasticity aspects, where plasticity indicates the network's ability to learn novel information, and stability reflects its ability to retain previously learned knowledge~\cite{wickramasinghe2023continual}. Broadly, continual learning can be evaluated in task-incremental, domain-incremental, and class-incremental learning scenarios. In this work, we emphasize the domain-incremental learning scenario, in which the system is expected to learn multiple tasks without providing their identity and incorporating explicit boundaries~\cite{mirza2022efficient}. 

There are various approaches that have been proposed in the literature to enable continual learning without experiencing catastrophic forgetting. These approaches can be classified into: regularization-based method, dynamic architecture, and complementary memory system (also known as replay or rehearsal)~\cite{wang2024comprehensive}. The regularization-based methods alleviate catastrophic forgetting through imposing constraints on the synaptic weight updates according to the task being learned, specifically identifying the critical synapses to the learned tasks and making it more rigid to future updates. This method is known to be computationally and memory efficient, but it is less effective compared to other methods~\cite{kwon2023lifelearner}. Examples include Elastic Weight Consolidation (EWC)~\cite{kirkpatrick2017overcoming}, Synaptic Intelligence (SI)~\cite{zenke2017continual}, and Learning without Forgetting (LwF)~\cite{li2017learning}. 
In dynamic architecture, the network dynamically changes its architecture via adding or removing neurons and synaptic pathways to accommodate new knowledge~\cite{piyasena2020dynamically}. Despite its promising performance, the dynamic architectures demand a reconfigurability feature and high computational cost. The complementary memory-based method involves fine-tuning network parameters through rehearsal, where previously learned information is revised during learning~\cite{van2020brain}. Rehearsing can be performed using a small subset of examples from previous tasks (experience replay)\cite{robins1995catastrophic}, stored neuronal activations of intermediate layers, or previous task examples generated using an additional model (generative replay)~\cite{shin2017continual}. 

The approaches mentioned above are effective for addressing catastrophic forgetting algorithmically, but they face additional challenges when realized in hardware, especially the mixed-signal memristive architecture. For instance, the regularization-based method involves storing additional information and repeatedly accessing parameters, incurring significant analog-to-digital conversions and increased memory traffic. The dynamic-architecture method demands a structural reconfigurability, which is not feasible in crossbar-based designs where connectivity is physically fixed. Besides that, both methods demand repeated and precise weight updates that challenge the endurance and stability of the memristor devices. These limitations make replay-based methods more appealing for hardware implementation as they avoid architectural expansion and reduce the need to compute second-order statistics and frequent memristor updates. Therefore, the proposed M2RU has complementary memory-based mechanism, particularly, the experience replay, as it is known to be the most effective approach especially when dealing with time-series data~\cite{sodhani2019toward}. All samples used during replay are selected using a reservoir sampler due to the temporal nature of input streams, i.e. non-stationary and unknown length. More details on MiRU replay are provided in~\cite{zyarah2025minion}.



\subsection{Minion Recurrent Unit (MiRU)}
\begin{figure*}[!t]
    \centering
    \includegraphics[width=0.9\textwidth]{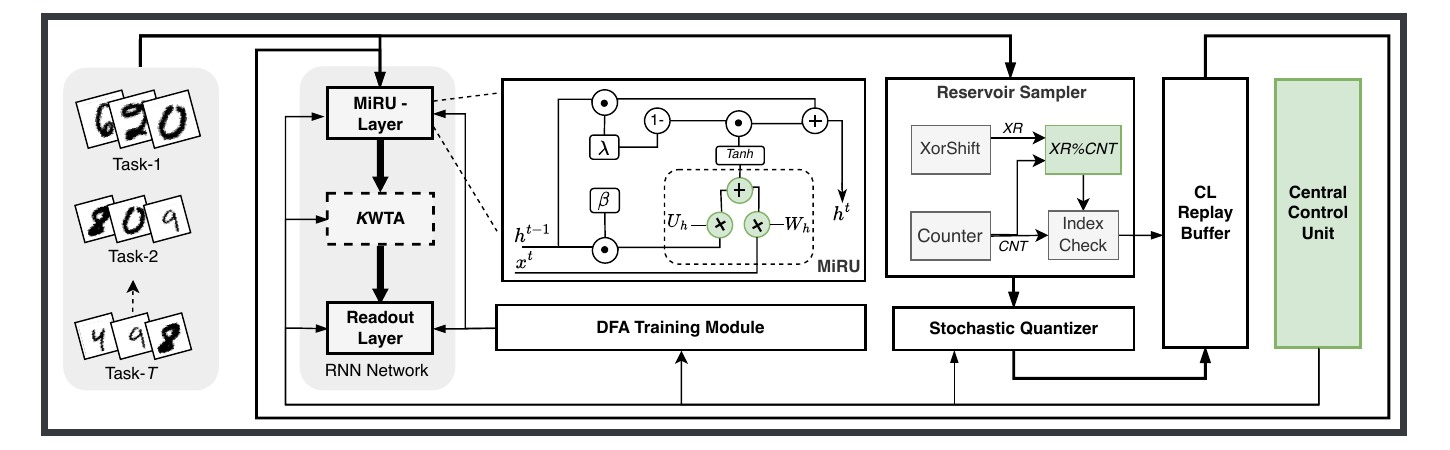}
    \caption{High-level block diagram of the proposed memristor-based MiRU accelerator, which consists of an RNN network to process time-series information, and a data preparation unit to randomly capture examples from non-stationary input streams to be stored in the replay buffer after quantization.}
    \label{sys_miru}
\end{figure*}

The Minion Recurrent Unit (MiRU) is introduced as a lightweight variant of the GRU~\cite{zyarah2025minion}. Despite its simplified structure, it is capable of adaptively capturing both short- and long-term temporal dependencies and has demonstrated performance comparable to GRU across various real-world tasks, including image classification and sentiment analysis~\cite{zyarah2025minion}. Similar to conventional recurrent networks, the MiRU RNN consists of three primary layers: input, hidden (MiRU layer), and output (readout), illustrated in~\fig{sys_miru}. These layers are interconnected through synaptic weights originating from either the preceding layer (forward connections) or the same layer at the previous time step (recurrent connections). Depending on the layer type and its role in the network, each layer performs a distinct function. The input layer, composed of linear neurons, acts mainly as a buffer that projects input features into the hidden layer through forward-weighted synaptic connections. Within the hidden layer, the MiRU units extract the relevant temporal features. Each MiRU receives two types of stimulation: feedforward input activations ($X \in \mathbb{R}^{n_x \times n_T}$)\footnote{Inputs are assumed to have a fixed sequence length $n_T$.} and recurrent activations from the previous time step ($H^{t-1} \in \mathbb{R}^{n_h \times n_T}$). Here, $n_x$ represents the input dimension, $n_h$ the number of hidden units, and $n_T$ the temporal length of each input sequence.

Each MiRU unit is governed by two coefficients, reset and update, which regulate the information flow and local storage. These coefficients are hyperparameters selected to shape the learning behavior of MiRU. Smaller update coefficients $\lambda$ promote the incorporation of new information, whereas larger values encourage stronger reliance on past hidden states. Likewise, a larger reset coefficient $\beta$ favors retaining previous hidden states, while a smaller one encourages rapid adaptation to new inputs. With appropriate choices of $\lambda$ and $\beta$, these coefficients can mimic the functional behavior of the update and reset gates in a standard GRU, balancing memory preservation with adaptability.

When inputs are presented to MiRU, the reset coefficient modulates the influence of the previous hidden state via element-wise multiplication with $h^{t-1}$. This determines the amount of historical information to be discarded, particularly information irrelevant to future predictions. When $\beta$ approaches zero, the hidden activation becomes almost entirely dependent on the current input. After applying the reset coefficient, the candidate hidden activation $\tilde{h}^t$ is computed from both the modified history and the new input as shown in~\eq{h_tilde_MiRU_2}. Once the candidate state is obtained, the update coefficient decides how much of the previous hidden activation influences the new hidden state. This mechanism is crucial for maintaining long-term memory. The final hidden activation $h^t$ is produced according to~\eq{h_out_MiRU_2} and is then passed to the output layer through the synaptic matrix $W_y \in \mathbb{R}^{n_h \times n_y}$ to generate the output $\hat{y}^t$ as shown in~\eq{yhat}.

\begin{equation}
    \tilde{h}^t = tanh(W_h x ^ t + U_h (\beta \odot h^{t-1}) + b_h)
\label{h_tilde_MiRU_2}
\end{equation}
\begin{equation}
    h^t = \lambda \odot h^{t-1} + (1 - \lambda) \odot \tilde{h}^{t}
\label{h_out_MiRU_2}
\end{equation}
\begin{equation}
    \hat{y}^t = \sigma(W_y~ h^{t}) 
\label{yhat}
\end{equation}

It is important to mention here that the MiRU equations reveal a substantial reduction in the number of components and parameters compared to the standard GRU. This simplification not only enhances interpretability, facilitating a deeper understanding of recurrent learning dynamics, but also significantly reduces storage demands and computational overhead during both training and inference. As a result, MiRU is far more suitable for deployment on resource-limited edge platforms, addressing challenges related to adaptation and latency.

\section{Direct Feedback Alignment Through Time}
RNNs with gating mechanisms are typically trained with Backpropagation Through Time (BPTT). Although BPTT enables RNNs to demonstrate state-of-the-art performance in numerous applications, including speech tasks~\cite{chen2024exploiting}, anomaly detection~\cite{ullah2022design}, and medical diagnosis~\cite{al2019prediction}, it imposes strict temporal dependencies that require storing every intermediate activation, which quickly overwhelms on-chip memory resources for long sequences. This is attributed to two factors: i)  transposed weights during error propagation, ii) backward locking problem, where the error should be transferred sequentially layer-by-layer~\cite{jot2020fefet}. To avoid these problems, several alternative training algorithms have been presented in the literature. Among them, Direct Feedback Alignment (DFA), which uses random and fixed weights to project the error to early layers in the network rather than using transposed forward weights. However, most DFA-based methods have struggled to match the performance of traditional BP when applied to recurrent architectures~\cite{han2020extension}.

In this work, we integrate DFA with gradient descent to enable efficient on-chip training of MiRU-based RNNs, see Algorithm~\ref{alg:miru-dfa}. The training begins by computing the gradient of the loss with respect to the output layer weights (Line 10) and projecting the loss to the earlier layers via the random weight matrix, $\psi$ (Line 13). Then, the gradient with respect to the hidden layer weights is computed as given in Lines 15-16. Unlike the output layer, here the gradient is accumulated back in time to capture long- and short-term dependencies. Finally, all the weights are updated according to gradient descent rules after being sparsified using the $K$-WTA function ($\zeta$), see Lines 19-21.

\IncMargin{1em}
\begin{algorithm}[t]
\caption{MiRU Training with DFA}
\label{alg:miru-dfa}
\KwIn{
 $\{(x^t,y^t)\}_{t=1}^{n_T}$, 
  $x^t \in \mathbb{R}^{n_x}$, $y^t \in \mathbb{R}^{n_y}$
}
\KwOut{Updated parameters $(W_o, W_h, U_h)$}
\DontPrintSemicolon

\textbf{Forward pass:}\\
\For{$t=1,\dots,n_T$}{%
  $\tilde{h}^t \leftarrow g(x^t W_h + (\beta\, h^{t-1}) U_h + b_h$)\;
  $h^t \leftarrow \lambda h^{t-1} + (1 - \lambda) \, \tilde{h}^{t-1}$\;
  $\hat{y}^t \leftarrow \sigma(h^t W_o + b_o$) \tcp*{$\sigma$ is the softmax}
}

Compute loss $ \ell(\hat{y}^t , y^t)$\;

\BlankLine
\textbf{Output layer gradients:}\\
$\delta_o^t \leftarrow \partial \ell / \partial (h^t W_o + b_o)$\;
$\nabla W_o \leftarrow (h^t)^\top \delta_o^t$ \;

\BlankLine
\textbf{Hidden layer:}\\
\For{$t=n_T,\dots,1$}{%
  $e^t \leftarrow \delta_o^t \Psi$\;
  $\delta_h^t \leftarrow \lambda\, e^t \odot g'\big(x^t W_h + (\beta\, h^{t-1}) U_h + b_h\big)$\;
  $\nabla W_h \mathrel{+}= (x^t)^\top \delta_h^t$ \;
  $\nabla U_h \mathrel{+}= (\beta\, h^{t-1})^\top z^t$ \;
}

\BlankLine
\textbf{Parameter update:} \\
$W_o + \leftarrow - lr \cdot \zeta (\nabla W_o $) \\
$W_h + \leftarrow - lr \cdot \zeta (\nabla W_h$) \\ $U_h +\leftarrow - lr \cdot \zeta (\nabla U_h$)
\end{algorithm}
\DecMargin{1em}

\section{System Design and Implementation}
Designing RNN hardware using hybrid memristor-CMOS technology requires careful consideration of data representation, movement, storage, and manipulation, as these aspects directly influence the interconnect network density, computational complexity, throughput, and overall energy efficiency.~\fig{sys_miru} illustrates the high-level block diagram of the proposed M2RU accelerator, which comprises a data preparation unit, MiRU-based RNN, and central control unit. In the following subsections, a detailed description of the hardware implementation of the core units of the accelerator is provided, while considering the aforementioned design aspects, aiming to enable efficient temporal data processing on edge devices with stringent resources.

\subsection{Data Preparation Unit}
The data preparation unit is used to randomly capture examples from non-stationary input streams to be stored in the replay buffer after processing. It comprises a reservoir sampler, a stochastic quantizer, and a replay buffer. A detailed description of each unit is provided in the following subsections:

\subsubsection{Reservoir Sampler}
The reservoir sampler randomly selects examples from non-stationary input streams of unknown length to be stored in the memory buffer. The selection of samples is done with continual updates and equal probabilities, making efficient use of the limited memory buffer. 
The process starts by filling the buffer of length $k$ with the first set of examples presented to the network. Then, for each new example position at $i$, where $i$ starts from $k+1$, a random integer number $j$ located between 1 and $i$ is generated. The $j^{th}$ number represents the index of the previous examples located within the buffer to be replaced with the new $i^{th}$ example, assuming $j$ is less than $k$. 

Implementing the reservoir sampler in hardware requires a variable-length random number generator, a counter, and index checker. The random number generator is used to generate random numbers  between 1 and the number of the presented examples, $i$ as counted by the counter. The index checker inspects if the generated random number $j$ falls within the size of the replay buffer $k$ to perform prospective overwrite. However, implementing a random number generator of variable length demands a reconfigurability feature which is costly to incorporate in hardware. To address this issue, we design the sampler with a modulus unit and 32-bit xorshift circuit, see~\fig{sys_miru}. When an input is presented to the network, the counter is increased and a random number is generated by the xorshift circuit. Since the xorshift range is between 1-$2^{32}$, the output may fall out of the counter range, 1-$i$. Thus, we used a modulus unit that computes the remainder of random numbers generated by the xorshift and the counter to generate a new random number (remainder) such that it always falls within the desired range. The xorshift here is utilized instead of a Linear Feedback Shift Register (LFSR) to ensure that every element in the presented temporal data has equal chances to be selected for replay, as xorshift produces decorrelated, uniform, and unbiased random indices, unlike LFSR.

\begin{figure*}[t!]
    \centering
    \includegraphics[width=1\textwidth]{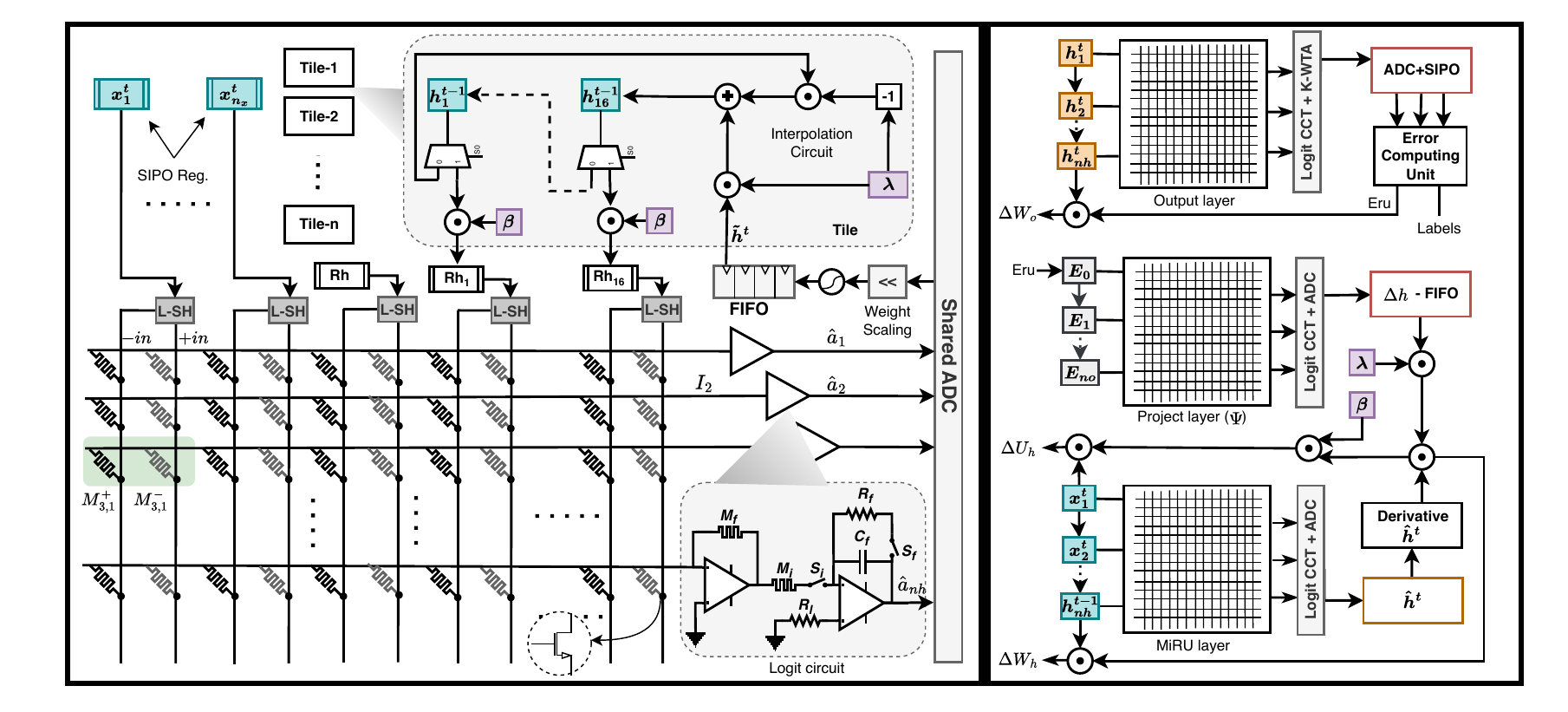}
    \caption{Left: Mixed-signal architecture of the memristive M2RU accelerator designed for processing temporally structured data. Right: High-level overview of the on-chip training framework that enables continuous network adaptation within dynamic environments.}
    \label{reset}
\end{figure*}

\subsubsection{Stochastic Quantizer}
This circuit is used to stochastically quantize input features stored in the replay buffer. It bridges the reservoir sampler and the replay buffer. As discussed earlier, the reservoir sampler continuously fills and updates the replay buffer with new examples. However, storing these selected examples in the replay buffer requires a non-trivial storage. To mitigate this challenge, all features are compressed using a lightweight quantizer to reduce the memory requirement by 2$\times$. This is accomplished via quantizing each feature (pixel) from 8-bit to 4-bit precision using stochastic rounding. This approach ensures unbiased quantization and reduces banding artifacts compared to plain truncation. Given an input pixel $x$ that needs to be quantized to an $n_b$ bits, the quantized value ($q$) is computed as: 
\begin{equation}
\label{scale}
z = x \cdot 2^{n_b},
\end{equation}
\begin{equation}
q =
\begin{cases}
\lfloor z \rfloor + 1, 
& \text{if } r < f_L \ \text{and}\ \lfloor z \rfloor < 2^{n_b}-1, \\[4pt]
\lfloor z \rfloor, 
& \text{otherwise}.
\end{cases}
\label{eq:scaled_stochastic_rounding}
\end{equation}
where, 
\begin{equation}
f_L = z - \lfloor z \rfloor, \qquad 
r \sim \mathcal{U}(0,1),
\end{equation}

In hardware, the Verilog model of stochastic quantizer can be implemented using a shift operator to scale the input features. This is followed by a fractional extraction circuit, an $n_b$ LFSR, comparator, and adder to apply the stochastic rounding rules.


\subsection{MiRU Recurrent Network}
The high-level diagram of the MiRU RNN is illustrated in~\fig{sys_miru}. It comprises three layers: input, hidden, and output. The hardware implementation of each layer is detailed in the following subsections:

\subsubsection{Hidden Layer}
The hidden layer comprises several MiRUs that mediate the input and output layers, and they are responsible for feature extraction and forming dependencies. As alluded to earlier, there are no explicit gates in MiRUs; rather, there are scaling factors controlling the retention and the flow of information. Therefore, the hidden layer can be implemented using a memristive crossbar, local buffers, neuron circuits, and ADCs. The memristive crossbar integrates the memristor devices emulating the synaptic weight in a compact form, enabling the parallel computation of vector-matrix multiplication (VMM) in the analog domain. Typically, each synaptic weight is mapped to two memristors, with their net conductance difference representing a bipolar weight. Given $n_x \times n_h$ connected layers with $n_h^2$ recurrent connections, their synaptic weights can be mapped to a crossbar structure that consists of $2\times [(n_x + n_h) \times n_h]$ array of tunable memristors and a fixed set of reference memristors (colored in grey) initialized at the midpoint of the resistance window~\cite{zyarah2018semi}. The net difference between any tunable memristor and the adjacent fixed memristor results in a bipolar weight.

\begin{figure*}[h!t]
    \centering
    \includegraphics[width=0.9\textwidth]{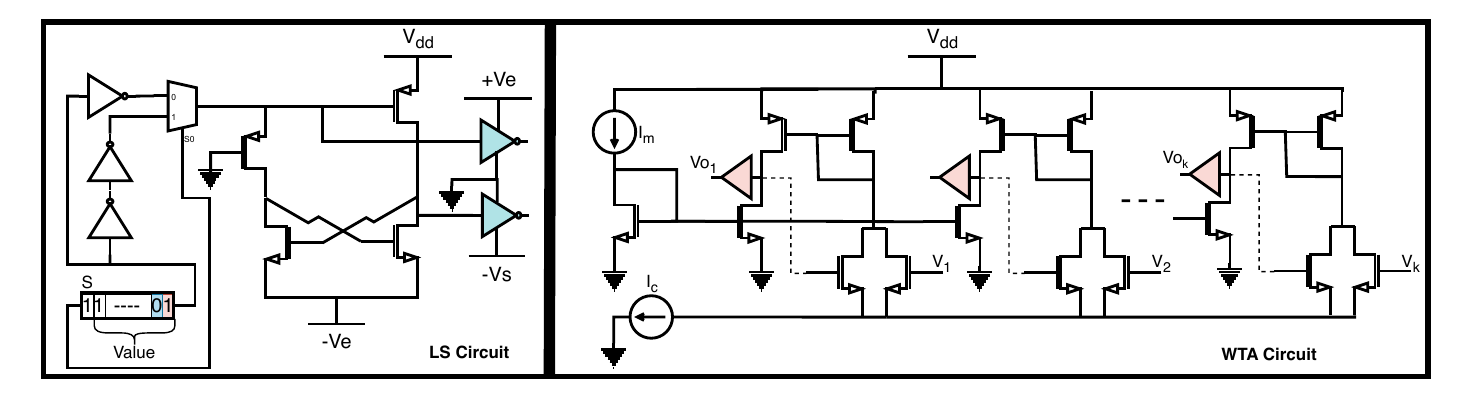}
    \caption{Left: Level-shifting circuit used to generate low-amplitude input pulses for weighted-bit streaming, supporting both positive and negative voltages for signed digital inputs. Right: $K$-WTA circuit used to approximate the softmax operation and sparsify gradients during on-chip learning.}
    \label{wta}
\end{figure*}

When an input feature vector $x^t$ (voltage) along with hidden layer activations from previous time step $h^{t-1}$ (voltage) scaled by $\beta$\footnote{One register is used to store the value of $\beta$ and it is shared across all MiRU units to minimize resources.} are presented to the crossbar wordlines, they are initially stored in local buffers and later fetched when required. The presentation of features and activations can be either in the analog or the digital domain. The former enables in-memory computing without data conversion, but requires sample-and-hold circuits to temporarily hold the activations until used in the current and subsequent time steps. Furthermore, it incurs challenges associated with: i) undesired leakage in the sample-and-hold circuit which leads to a drop in the stored charge, ii) unreliable intra- and inter-layer data movement, and iii) high resource utilization~\cite{zyarah2024time}. The latter considers digitally quantizing the input features and activations using fixed-point representation. This facilitates data movement, ensures signal integrity, enhances network performance, and allows robust and efficient integration with external sensors. While using digital representation instead of analog imposes the need for high-precision DACs and ADCs, this issue is addressed through resource sharing and the use of weighted-bit streaming, which allows the digitized input features to be presented in a sequential fashion, bit-by-bit (further details are provided in the methodology section). Whenever the input bits from the buffers are fetched to the crossbar, an output current is produced at each bitline, indicating the partial weighted-sum of the dot-product between the presented partial features ($k^{th}$ bits) and the corresponding weight matrix, see~\eq{neo_current}. Here, $M_{ji}$ refers to the resistance of the memristor connecting the $j^{th}$ output neuron with the $i^{th}$ input neuron, and $M_{ri}$ is the reference memristor resistance of the $i^{th}$ column.
\begin{equation}
\label{neo_current}
\begin{aligned}
     I_{j,k} = x^t_{1,k} \underbrace{( \frac{1}{M_{j,1}} - \frac{1}{M_{r1}})}_{\text{unscaled weight, $W_{j1}$}}... +x^t_{n_x,k}(\frac{1}{M_{j,n_x}}-\frac{1}{M_{r,n_x}}) + \\ h^t_{1,k} ( \frac{1}{M_{j,n_x +1}} - \frac{1}{M_{r,n_x+1}})... +h^t_{n_h,k}(\frac{1}{M_{j,n_h}}-\frac{1}{M_{r,n_h}})
\end{aligned}
\end{equation}

The bitline current is then integrated by the corresponding neuron circuit to generate logits of each neuron, $\hat{a}$. Each neuron circuit is composed of an inverting operational amplifier (Op-Amp) to provide a virtual ground to the crossbar bitlines and also to contribute in determining each input bit significance as they are presented sequentially. The inverting Op-Amp is followed by an integrator to perform analog accumulation of the partial weighted-sum current. Once the integration process is finished, the ADC converts the accumulated values to digital, which is then subjected to a non-linear activation function ($tanh$) after being scaled using a shift operation controlling the dynamic range of the synaptic weights. The output of the activation function is stored in a FIFO to be used in the subsequent time step.

It is important to mention here that due to the high power consumption of the ADCs, it is common to use multiple ADCs time-multiplexed across several bitlines~\cite{shafiee2016isaac, xu2024reharvest}. However, this requires the integrator to hold the accumulated charges long enough without major leakage. To quantify the retention requirements of the integrator during ADC scanning, the leakage behavior is modeled using \eq{leakage}, where $\Delta V$ represents the change in the stored voltage, $V_{int}$ or $\hat{a}$, $T_{conv}$ is the time required by the ADC to scan all channels, and $\tau$ is the time constant of the integrator controlled by $R$ and $C_f$. Since $V_{int}$ is restricted by the used technology and bias voltage, to minimize $\Delta V$ to less than 0.2LSB, one either has to use a large capacitor and resistor or multiple ADCs shared by a small number of bit-lines. However, both these solutions turn out to be costly in terms of resources and energy. Thus, to circumvent this challenge, we insert transmission gates (switches) at the summing node ($S_i$) and at the feedback of the integrator ($S_f$). During the hold phase, both switches are opened, and the charge leakage will be limited to the Op-Amp input bias current ($I_b$) and the capacitor dielectric leakage, given by~\eq{leakage2}\footnote{Eq.~\eq{leakage} can be approximated to~\eq{leakage2} as $T_{conv}$ is much smaller than $\tau$.} and \eq{leakage3}. 

\begin{equation}
\label{leakage}
    \Delta V = V_{int} \times \exp({\frac{-T_{conv}}{\tau}})
\end{equation}
\begin{equation}
\label{leakage2}
    \Delta V_l \approx V_{int} \times \frac{T_{conv}}{R_{leakage}~ C_f}
\end{equation}
\begin{equation}
\label{leakage3}
    \Delta V_b = \frac{I_b ~T_{conv}}{C_f}
\end{equation}

Besides using switches, high-speed ADCs, similar to the work proposed by~\cite{shafiee2016isaac}, are utilized to keep the change in the accumulated voltage negligible over the scan interval. We used one high-speed ADC shared by all bitlines for a given crossbar. This ADC offers a sampling rate of 1.28 GSps ($T_{conv}$ per channel is $\sim$2 ns). Thus, the total $\Delta V$ is estimated to be less than 10.5 $\mu$v ($ < 0.1 LSB$) over 200 ns (worst case) with $C_f$ = 2 pF, $I_b$ less than 50 pA, and $R_{leakage} > 10~$G$\Omega$. 

The output of the ADC, which is subjected to non-linearity and stored in the FIFO, represents the candidate hidden states ($\tilde{h}_t$) of the MiRU units. Each candidate hidden state should be multiplied by $\lambda$ and added to the corresponding activation from the previous time step scaled by 1-$\lambda$ (linear interpolation) to compute the final output of MiRU units. Performing such operation for all MiRU units concurrently is a costly process as it demands numerous multipliers, adders, etc. To cut down the resources, this work suggests computing the hidden state activations in a hybrid form (concurrent and sequential\footnote{The sequential computation of the hidden layer activations is not possible or extremely challenging in the analog domain.}) to balance the resource usage and network throughput. The hidden layer is divided into several tiles that work concurrently at the layer level and sequentially within the tile. For a given tile, at each time step, the output of one MiRU unit is computed. This is performed by fetching a candidate hidden state and loading the corresponding $h^{t-1}$ from the shift registers with a configurable dataflow\footnote{The shift register works in Serial-in-Parallel-out (SIPO) mode during candidate hidden-state computation and in Serial-in-Serial-out (SISO) mode otherwise.}. Once the final outputs of the MiRU units are computed, they replace the previous activations stored in the shift register and are presented to the subsequent layer, the $k$-WTA\footnote{Detailed description of the $k$-WTA operation can be found in~\cite{zyarah2025voltage}.} (when used) or readout layer.

\subsubsection{Readout Layer and Training Module}
The readout layer has an analogous architecture to that in the hidden layer, yet incorporates only forward synaptic connections projecting the hidden-layer activations to the output neurons. These connections are realized with memristor devices integrated into a crossbar structure, enabling efficient accumulation of the partial weighted-sum in the integrator. The integrator is followed by a voltage-mode $k$-winner-take-all circuit ($k$-WTA), shown in~\fig{wta}-Right, to apply the softmax function. The $k$-WTA output is subsequently digitized by an ADC and used to evaluate the network’s prediction accuracy during both inference and training.

Performing the training locally on-chip in a memristor-based architecture encounters several challenges, primarily due to the overhead associated with resource utilization, data movement across mixed-signal interfaces, and temporal storage of intermediate results. In this work, these challenges are mitigated through efficient data-conversion strategies and integration of DFA and gradient descent optimization. The training stage begins by presenting the ground-truth labels to the readout layer and estimating the prediction error, which represents the net difference between the labels and the network output. This is performed using the error computing unit, whose output ($E_{ru}$) is sequentially propagated to both the output and hidden layers. 

In the output layer, the error is multiplied by hidden-layer activations to form the weight update term $\Delta W_o$, see~\fig{reset}-Right. These updates are used to modulate the conductance of the corresponding memristors using the Ziksa programming scheme~\cite{zyarah2017ziksa}. To reduce memory overhead and prevent unintended perturbation of output synapses, only the hidden activation corresponding to the current input sequence $x^{n_T}$ is used to compute $\Delta W_o$. Unlike the output layer, training the hidden layer is more demanding because it requires access to all input features presented over the last $n_T$ time steps, in addition to the corresponding hidden activation. To avoid storing the entire sequences, the input features are stored in auxiliary memory, while hidden activations are recomputed on demand as in the inference stage. This process begins by projecting $E_{ru}$ through a projection layer of random and untuned weights ($\Psi$), consistent with the DFA learning rule. The resulting projected error is stored in a FIFO buffer and sequentially used to compute the gradients associated with the input weights $W_h$ and recurrent weights $U_h$. These gradients are then translated into conductance modifications and written to the memristor array.


\begin{figure*}[th!]
\centering
\subfigure{\includegraphics[width = 0.24\textwidth]{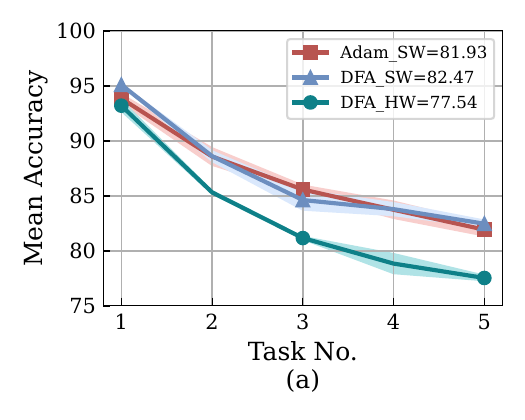}}
\subfigure{\includegraphics[width = 0.24\textwidth]{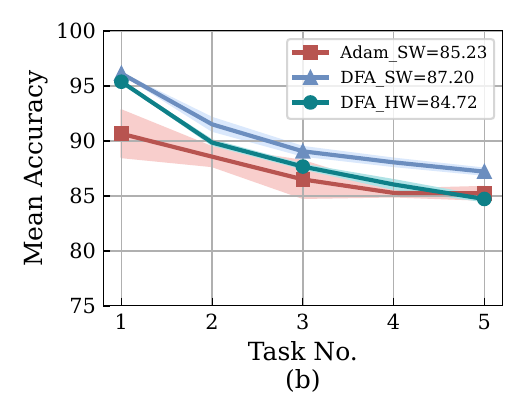}}
\subfigure{\includegraphics[width = 0.24\textwidth]{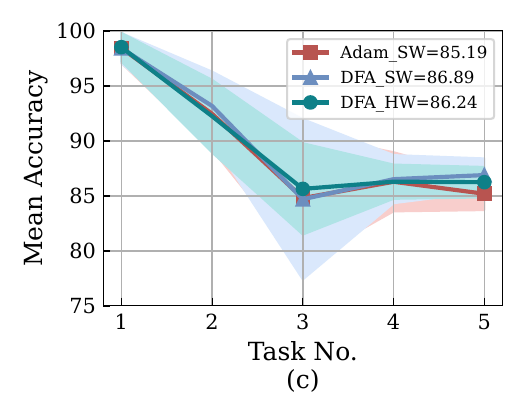}}
\subfigure{\includegraphics[width = 0.24\textwidth]{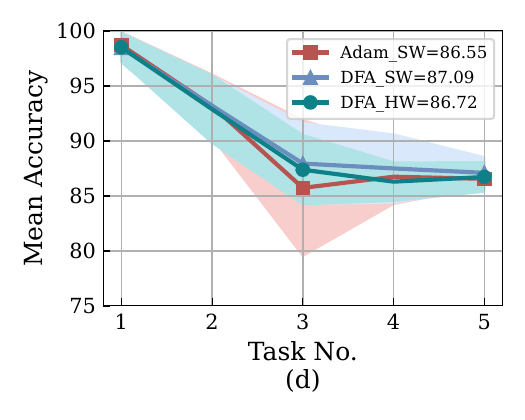}}
\caption{Average test accuracy (after each task) of the proposed M2RU accelerator along with software counterpart trained with DFA and Adam optimizer when verified on sequential tasks from permuted MNIST (a: $n_h$ = 100, b: $n_h$=256) and split CIFAR-10 (c: $n_h$ = 100, d: $n_h$=256) datasets.}
\label{accur_thx}
\end{figure*}

\section{Methodology}
This section presents the methodology used to realize partial computation in the proposed mixed-signal M2RU accelerator using weighted-bit streaming (WBS). Details regarding the physical characteristics of the memristor devices, their programming schemes, and the experience-replay setup can be found in our prior work~\cite{zyarah2024time, zyarah2025minion}.

\subsection{Weighted-Bit Streaming} 
Processing analog signals in mixed-signal neuromorphic systems is often constrained by the high overhead caused by data conversion, temporary storage, and intra- and inter-layer data movement. Such problems typically dominate recurrent neural networks due to their feedback. To address this issue, this work presents Weighted-Bit Streaming (WBS) as a partial computing strategy, enabling multi-bit digital inputs to be processed efficiently within the memristive crossbar without requiring high-resolution DACs, high-precision conversions, or duplicated datapaths.

The key idea behind WBS is that the digital features or activations are streamed serially, one bit at a time. Each bit is multiplied by the corresponding memristive synaptic weight and accumulated on the bitline through Kirchhoff’s current law. Directly streaming all bits with equal significance would distort the final dot-product result. Prior implementations addressed this by encoding bit significance using exponentially increasing pulse durations~\cite{marinella2018multiscale}, but such exponential timing introduces substantial latency and is less attractive to time-critical applications. 

To overcome this limitation, we introduce a memristor-based bit-significance scaling mechanism, in which each streamed bit is weighted through an analog gain rather than through time encoding. One could use one memristor ($M_f$) placed in the feedback circuit of the sensing amplifier to achieve an absolute gain of $\frac{M_f}{M{ji}}$, where $M{ji}$ is the synaptic device resistance. However, for a digital input normalized between [0,1] quantized with $n_b$ bits, $M_f$ should span [$2^{-1}$ to $2^{-{n_b}}] \times M_{ji}(min)$, corresponding to a range of more than two orders of magnitude if $n_b$ is 8-bit. Given the maturity of the memristor devices, such a range is not practical. To circumvent this limitation, we replace the fixed input resistor of the integrator with another tunable memristor, ${M_{i}}$. This enables the bit significance to be determined by the ratio $\frac{M_f}{M_{i}}$, see~\fig{reset}-Left. This ratio-based tuning substantially reduces the required resistance window  and makes the implementation physically feasible.

Given that all input bits are presented using WBS, the output of the integrator is expressed as a weighted sum of the partial contributions:

\begin{equation}
\label{integrator}
    V_{int} = \frac{-1}{M_i C_f} \Big[ \int_{t_0}^{t_1} V_{x,1}~dt + .... + \int_{t_{n_b - 1}}^{t_{n_b}} V_{x,{n_b}}~dt \Big]
\end{equation}

Since each input bit indexed by $k$ is presented as a pulse of fixed duration ($T_s$), 
\begin{equation}
    \int_{t_{k-1}}^{t_k} V_{x,k}~dt = V_{x,k} \times T_s
\end{equation}
The integrated voltage becomes
\begin{equation}
    V_{int} = \frac{-1}{M_i C_f} \Big[ V_{x,1}\times T_s + .... +  V_{x,{n_b}}\times T_s \Big]
\end{equation}
\begin{equation}
    V_{int} = \frac{T_s}{M_i C_f} \Big[ M_{f,1} \times I_{x,1} + .... + M_{f,n_b} \times I_{x,{n_b}} \Big]
\end{equation}
When both $M_{i}$ and $M_f$ are tunable, this generalizes to: 
\begin{equation}
\label{xyz}
    V_{int} = \frac{T_s}{C_f} \sum^{n_b}_{k=1} (\frac{M_f}{M_{i}})_k \times I_{x,k} 
\end{equation}
To avoid saturating the integrator before presenting all input bits to the network, the capacitor of the integrator should be estimated based on the worst-case scenario, where each bit produces the maximum bitline current $I_{max}$, the expression becomes:
\begin{equation}
    V_{int} = \frac{I_{max} \times T_s}{C_f} \sum^{n_b}_{k=1} (\frac{M_f}{M_{i}})_k
\end{equation}
Given bit significance implemented via memristor ratios,
\begin{equation}
    \sum^{n_b}_{k=1} (\frac{M_f}{M_{i}})_k = \sum^{n_b}_{k=1} 2^{-k}
\end{equation}
According to finite geometric series, 
\begin{equation}
    \sum^{n_b}_{k=1} 2^{-k} = 1 - 2^{-nb}
\end{equation}
Therefore, the integrator voltage can be approximated as in \eq{final_eq}, where the $C_f$ value can be selected based on the desired output swing. Here, the $C_f$ value is set to be 1 $pf$, given the total $I_{max}$ per bitline (worst case) is $\approx$3.2$\mu A$ and $T_S = 50 ns$.
\begin{equation}
\label{final_eq}
    V_{int} \approx \frac{I_{max} \times T_s}{C_f} 
\end{equation}

To reduce the capacitor area and maintain energy efficiency, all the digital bits presented to the network are level-shifted to a low amplitude of 0.1v using the modified level-shifter in~\fig{wta}-Left, originally proposed by~\cite{dame1974voltage}. The circuit also supports signed inputs, where a digital “1” is streamed as either a positive or negative voltage depending on the sign of the encoded value, and “0” is streamed as 0 V. Compared with prior architectures such as ISAAC~\cite{shafiee2016isaac} and PANTHER~\cite{ankit2020panther}, the proposed partial-computing method avoids the need for high-precision DACs, does not require crossbar or column duplication to support multi-bit inputs. Furthermore, it assigns bit significance through memristor-ratio scaling rather than time-encoded pulse widths or digital shift-and-add operations, enabling a more compact and energy-efficient mixed-signal architecture.

\subsection{Experimental Setup}
In order to assess the performance of the proposed M2RU prior to hardware implementation, we employ a combination of software models, hardware-like mixed-signal model, and full custom hardware simulation. The software models are developed in PyTorch and used primarily for design space exploration while considering two learning algorithms: i) Backpropagation with the Adam optimizer, and ii) DFA, which is more suitable for hardware-constrained learning. The software models also enable rapid estimation of network accuracy and convergence behavior. 

The hardware-like model is utilized to emulate device- and circuit-level non-idealities. It incorporates the effects of various design constraints, such as memristor non-linearity and variability, quantization noise, and limited biasing voltage. The hardware-like model allows full network replication and parallel execution to reduce simulation time while preserving fidelity to mixed-signal operation. For the hardware implementation, all digital units are modeled in Verilog HDL and physically synthesized using Cadence-Genus. The resulting gate-level netlist is then imported into Cadence-Virtuoso, where it is integrated with the analog modules that implement the core computational blocks of M2RU. The complete mixed-signal design is simulated to precisely measure the latency, throughput, and average power consumption. For the memristor devices, to ensure practicality, we used a Verilog-a memristor model~\cite{kvatinsky2015vteam} fitted to the physical device characteristics proposed in~\cite{yang2010high}. To comply with the device characteristics, the used technology node, and targeted application constraints, the following has been considered: i) the memristor offers a high conductance range ($R_{on}$=2 M$\Omega$ and $R_{off}$=20 M$\Omega$), ii) the set and reset voltage is no higher than 1.2 v, where as the device threshold is set to $\pm$1 V,  iii) the device variability is limited to 10\%\footnote{Memristor device non-idealities considered during the simulation are: 10\% cycle-to-cycle variability (memristor resistance) and device-to- device variability (write variation).}.

\begin{figure*}[th!]
\centering
\subfigure{\includegraphics[width = 0.24\textwidth]{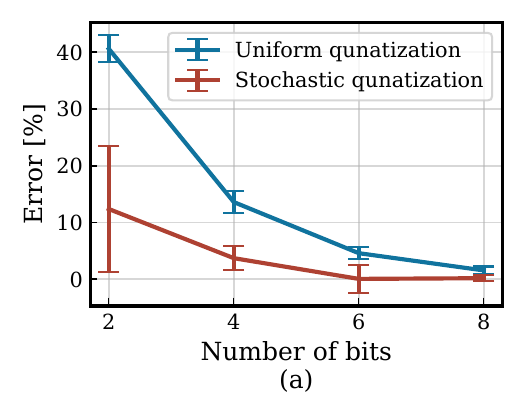}}
\subfigure{\includegraphics[width = 0.24\textwidth]{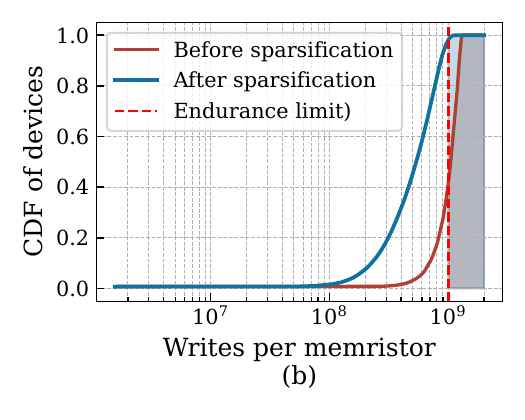}}
\subfigure{\includegraphics[width = 0.24\textwidth]{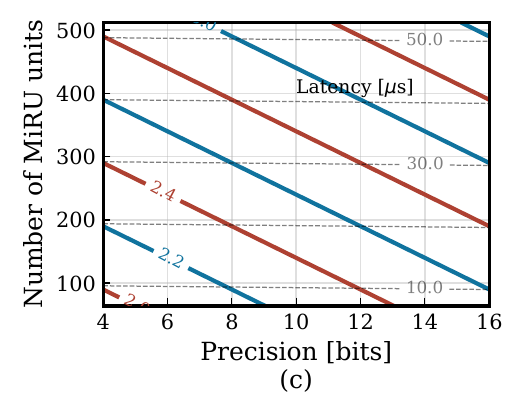}}
\subfigure{\includegraphics[width = 0.24\textwidth]{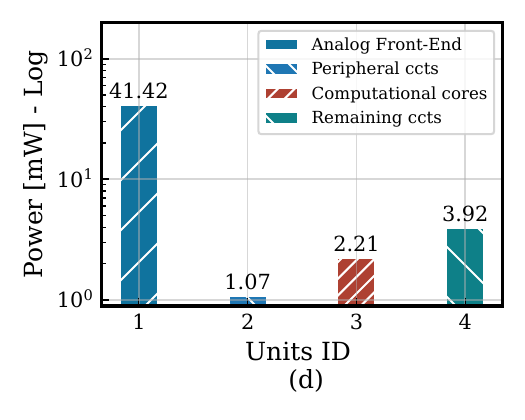}}
\caption{(a) The average percentage error when performing Matrix-vector multiplication during replay under uniform and stochastic quantization, (b) Estimated lifespan of M2RU before and after applying gradient sparsification, (c) Impact of network scaling and bit-precision on network latency with and without tiling (dotted lines), and (d) Breakdown of power consumption across the core units of the M2RU accelerator.}
\label{metric}
\end{figure*}
\section{Experimental Results and Analysis}

\subsection{Continual Learning and Catastrophic Forgetting}
The performance of the proposed memristor-based accelerator is quantified using a sequence of continual learning tasks derived from permuted MNIST and split CIFAR-10\footnote{Although MNIST and CIFAR-10 are image datasets, they are used here as sequential task streams to evaluate temporal continual learning behavior.}. Both the hardware model and the software models trained with the Adam optimizer and DFA are assessed under a domain-incremental learning scenario, in which all tasks share the same output layer and no task identity is provided during inference.~\fig{accur_thx} reports the average test accuracy (after each task) for networks with 100 and 256 MiRU units in the hidden layer. The mean accuracy ($MA$) after learning all tasks is computed using~\eq{mean_acc}, where $R_{T,i}$ denotes the accuracy of the model on task $i$ after learning all $T$ tasks sequentially\footnote{Details about the replay setup with MiRU are provided in~\cite{zyarah2025minion}.}.
\begin{equation}
    MA = \frac{1}{T}\sum\limits^T_{i=1} R_{T,i}
\label{mean_acc}
\end{equation}

Across all models, the network initially demonstrates high accuracy on early tasks and subsequently begins to experience graceful degradation in performance as new tasks are introduced. Despite this degradation, catastrophic forgetting is effectively avoided due to the use of replay buffers. However, the degradation tends to be steeper when using 100 MiRU units due to the limited learning capacity and increased overlap in neuronal activities dedicated to learning different tasks. When comparing software models, it can be observed that a $\sim$(1-2)\% improvement in mean accuracy is achieved when training the network using DFA compared to Adam optimizer. This gap is widened compared to the hardware counterpart and is estimated to be 4.93\% on permuted MNIST, but can be reduced when the hidden layer size is expanded to 256 neurons to become 2.48\%. For CIFAR-10, the mean accuracy remains almost similar across models as all features are extracted using pre-trained ResNet-18 from Torch library. However, larger variations are observed, which can be  attributed to two reasons, the first being the limited number of examples used in each task ($\sim$10,000 examples) compared to permuted MNIST. Second, the small size of the replay buffer dedicated to storing examples from previous tasks. The buffer size is experimentally set to store 1875 examples per task for permuted MNIST and 312 examples per task for CIFAR-10. All features stored in the replay buffer are stochastically quantized to reduce storage requirements. Unlike uniform quantization, stochastic quantization has a negligible impact on network performance, even when reduced to 4-bit, as the total error produced by vector-matrix multiplication of the features with the corresponding weights remains below $\sim$5\%, see~\fig{metric}-(a).

\subsection{Lifespan and Reliability}
The M2RU is designed with the aim of handling tasks of a continual nature in a domain-incremental scenario. However, the limited endurance of the memristor devices that endow the accelerator with its elasticity feature may hinder this goal. Memristor devices are fabricated with endurance ranging from $10^6 - 10^{12}$ switching cycles, which defines the number of permissible write operations before failure~\cite{kapur2024ultra}. Typically, during the training process, the memristors experience a certain degree of uniformity in write operations owing to the lack of sparsification in neuronal activities, especially when using differentiable bounded activations, as in gated RNNs. This eventually shortens the effective lifespan of the network and makes it less reliable for critical applications. 

To mitigate this limitation, gradient sparsification is adopted to reduce the number of write operations per memristor.~\fig{metric}-(b) illustrates the Cumulative Distribution Function (CDF) of memristor writes before and after applying gradient sparsification when classifying permuted MNIST, with the distributions projected forward to the endurance limit. The shaded regions beyond the endurance limit represent the overstressed memristors. Before sparsification, the CDF curve exhibits a sharp rise, reflecting the uniformity of write operations, with an average of 1.6$\times 10^{5}$ and an abrupt loss of elasticity feature for 58.28\% of the memristors. After sparsification\footnote{The sparsification ratio of gradient is set to $\sim$43\% without experiencing drop in performance. Further reduction, for example, to 30\%, can cause 3-4\% a drop in MA.}, the average of write operations drops to 8.5 $\times 10^4$, a reduction of $\sim$47\% in total write activity. In addition, the loss of elasticity becomes gradual, thereby extending the overall lifespan of learning. 

When translating the aforementioned numbers in time to project the lifespan in years, M2RU endows the user with $\sim$6.9 years of reliable usage without experiencing failure, assuming the learning is performed at a rate of 1 ms and that the memristor endurance is $10^9$ cycles. With gradient sparsification, the expected lifespan increases to about 12.2 years. One could also extend the lifespan even further, potentially by a factor of two, if frozen memristors are used for learning. It is imperative to mention here that these estimates may vary as the frequency of parameter updates is affected by the statistical activities of the input space~\cite{zyarah2020neuromorphic}.






\subsection{Throughput}
M2RU throughput is measured as the number of examples that can be processed sequentially per unit time. It is calculated to be $\sim$19,305 seq/s, enabled by an efficient hardware design with a throughput of $\sim$15 GOPS. This high throughput primarily stems from the high parallelism of performing the most dominant operation, MAC computation, as well as additional parallelism across layers and within each layer during data processing and movement. 

\fig{metric}-(c) illustrates the impact of network scaling and bit precision on processing latency. Increasing bit precision causes a linear increase in latency. This impact remains marginal when no tiling is applied in the hidden layer, as the dominant source of delay is the serialized interpolation of the candidate states in the MiRU units, see the dotted lines in~\fig{metric}-(c). In contrast, when tiling is used, the contribution of bit precision becomes more substantial, accounting for nearly one-third of the total delay. 

Regarding network scaling, increasing the number of MiRU units results in a noticeable increase in the delay. However, this effect drops significantly when tiling is applied. The tiling ensures that no more than 16 cycles are required to perform the interpolation of the MiRU units regardless of the hidden layer size. Given the network setup and applied tiling, the latency expected to process one set of features is 1.85 $\mu$s. This latency can be reduced either by boosting the system clock (set to 20 MHz) or increasing the degree of parallelism within the hidden layer even further through a larger number of tiles. However, this comes at the cost of duplicating the shared resources, such as the  linear interpolation circuit. To compromise between resource usage and throughput, we set the number of tiles to (4 - 16), depending on the adopted network architecture.


\begin{table*}[h!t]
\footnotesize
\caption{A comparison of the proposed RNN accelerator with previous memristor-based ASIC accelerators designed for classification tasks. One may note that these implementations are on different substrates, thereby this table offers a high-level reference template for RNN hardware rather than an absolute comparison.}
\label{HardwareAnalysis}
\begin{center}
\begin{threeparttable}
\begin{tabular}{|c|ccccc|}
\hline                      
\rowcolor{Gray} 
\textbf{Algorithm} & \textbf{M-GRU \cite{zhang2022memristor}} & \textbf{MDGN \cite{wang2024mdgn}} & \textbf{HGRU~\cite{pu2025memristor}} & \textbf{MBLSTM~\cite{liu2020memristor}} & \textbf{This work} \\ \hline 
  Operating Frequency  & - & 200 MHz &  -  & - & 20 MHz\\ 
  Network Size  & 6$\times$8k$\tnotex{tnote:robots-a4}\times$36 & 3$\times$150$\times$1 & 28$\times$128$\times$10 & - & 28$\times$100$\times$10\\ 
  Averaged consumed power & 173.65 mW\tnotex{tnote:robots-a2}  & 25.07 mW\tnotex{tnote:robots-a2} & - & $<$1.5W & 48.62 mW\\ 
  Dataset &	CASIA  & CALCE & MNIST \& IMDB & MNIST \& IMDB & MNIST \& CIFAR-10\\
  Latency  &  45 ns\tnotex{tnote:robots-a4}      &       1.22 s       &  5.14 $\mu$s  & - & 1.85 $\mu$s  \\ 
    RNN topology & GRU & GRU      &  Minimal GRU  & LSTM & MiRU  \\ 
  Technology node &  40 nm  & - &  -  & -  & 65nm   \\ 
   CL Learning & No & No & No & No & DIL-CL\\
  Training & Off-Chip & Off-Chip & Off-chip & On-Chip & On-Chip
 \\ \hline
  \end{tabular}
      \begin{tablenotes}
      \item\label{tnote:robots-a2} The work in~\cite{wang2024mdgn} reports the power for a MDGN (Memristive denoising recurrent unit) instead of the overall network.
    \item\label{tnote:robots-a4} The authors used 8-hidden layers of GRU, but did not mention the size. The execution time for each GRU cell is 45 ns and latency details are not provided. \vspace{-3mm}
    \end{tablenotes}
\end{threeparttable}
\end{center}
\end{table*}



\subsection{Power Consumption}
To estimate the power consumption of the proposed mixed-signal design of M2RU, all digital units are custom-built in Verilog HDL and physically synthesized using Cadence Genus. The generated netlist\footnote{Small-scale version of the design, 32x16x5, is verified for functionality to speed up the simulation.} is verified for functionality after being imported into Cadence Virtuoso and integrated with the analog blocks. The system is clocked at 20 MHz and simulated using Cadence ADE. The average power consumption is estimated to be $\sim$48.62 mW while predicting streaming data derived from the MNIST dataset. During the training process, the average power consumption increases to $\sim$56.97 mW due to the activation of additional circuitry, including the projection circuit to transfer network error and the write-control logic for weight updates. 

\fig{metric}-(d) illustrates the breakdown of the core units in the M2RU accelerator with $n_h$=100. It can be observed that most of the power is directed towards the analog front-end circuits, particularly the ADCs and Op-Amps used for neuron circuits. To mitigate this overhead, a high-speed ADC is shared among all neurons within a layer (for layer sizes $\le$ 128 neurons) and no activation function is used in the analog domain. Instead, to apply the non-linearity, a $tanh$ activation function modeled with a piecewise-linear digital implementation, which consumes solely $\sim$3.74 $\mu$W, is utilized. Aligning with the serialization of reading the weighted-sum of neurons, the activation function is shared across all neurons to reduce the overall power consumption. The sharing principles is also applied to scaling coefficients, reset and update, and to the interpolation of the candidate hidden states. Besides resource sharing, voltage-scaling of encoded digital bits presented to the memristor crossbars and gradient sparsification contribute significantly to enhancing the energy efficiency of the design, as it reduces the switching activity and the number of memristor write operations. This is further complemented by the weighted bit-streaming, which eliminates the need for high precision DACs to convert digital inputs to analog, especially when interfacing M2RU with computer systems.

\tb{HardwareAnalysis} provides a high-level comparison between M2RU and memristor-based accelerators in the literature. The overall energy efficiency of the M2RU accelerator is estimated to be 312 GOPS/W, corresponding to 3.21 pJ/op. Under the same technology, this represents a 29$\times$ improvement compared to the digital MiRU implementation. Such a gain demonstrates the potential of M2RU as a highly energy-efficient solution for edge applications with stringent power and resource constraints.

\section{Conclusions}

This work presented M2RU, a mixed-signal recurrent accelerator that enables continual learning through a compact Minion Recurrent Unit architecture and a hardware compatible training framework based on direct feedback alignment. A weighted bit streaming method was introduced to support multi bit input processing in memristive crossbars with low conversion overhead, and an integrated replay strategy was used to maintain stable adaptation under domain shift. 

The evaluation of catastrophic forgetting on sequential tasks shows the retention of previously learned tasks, with minor difference in performance between the model and mixed-signal architecture. 
In terms of computational performance, M2RU offers a throughput of $\sim$19,305 seq/s, enabled by efficient hardware design with a throughput of $\sim$15 GOPS. 
By combining weighted-bit streaming, shared ADC/DAC modules, and neuron circuits, the architecture achieves a high energy efficiency of 312 GOPS/W, representing a 29$\times$ improvement compared to its digital counterpart. The lifespan and reliability analysis demonstrate that M2RU can support $\sim$12.2 years of continual learning at 1 ms update rate, achieved through gradient sparsification.Overall, M2RU opens up new possibilities for edge AI devices that must function under tight resource limitations.



\ifCLASSOPTIONcaptionsoff
  \newpage
\fi
\bibliographystyle{IEEEtran}
\bibliography{IEEEabrv,References}


\begin{IEEEbiography}
[{\includegraphics[width=1in,height=1.25in,clip,keepaspectratio]{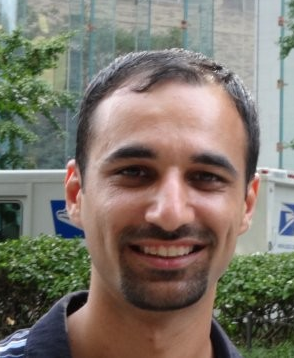}}]{Abdullah M. Zyarah} is an Assistant Professor in the Department of Electrical Engineering at University of Baghdad, and a Research Scientist in the Department of Electrical and Computer Engineering at the University of Texas at San Antonio. He specializes in digital and mixed-signal design, with research interests spanning neuromorphic architectures, memristive computing, energy-efficient machine learning, and biologically inspired learning algorithms. He previously worked at Seagate Technology, USA, as a research intern, where he contributed to neuromorphic hardware research and the development of mixed-signal circuits for brain-inspired computing. Dr. Zyarah received the B.Sc. degree in electrical engineering from the University of Baghdad, Iraq, in 2009, and the M.Sc. degree in the same discipline from Rochester Institute of Technology, USA, in 2015, where he also earned his Ph.D. degree in electrical and computer engineering in 2020. 
\end{IEEEbiography}

\vskip -2\baselineskip plus -1fil
\begin{IEEEbiography}
[{\includegraphics[width=1.25in,height=1.2in,clip,keepaspectratio]{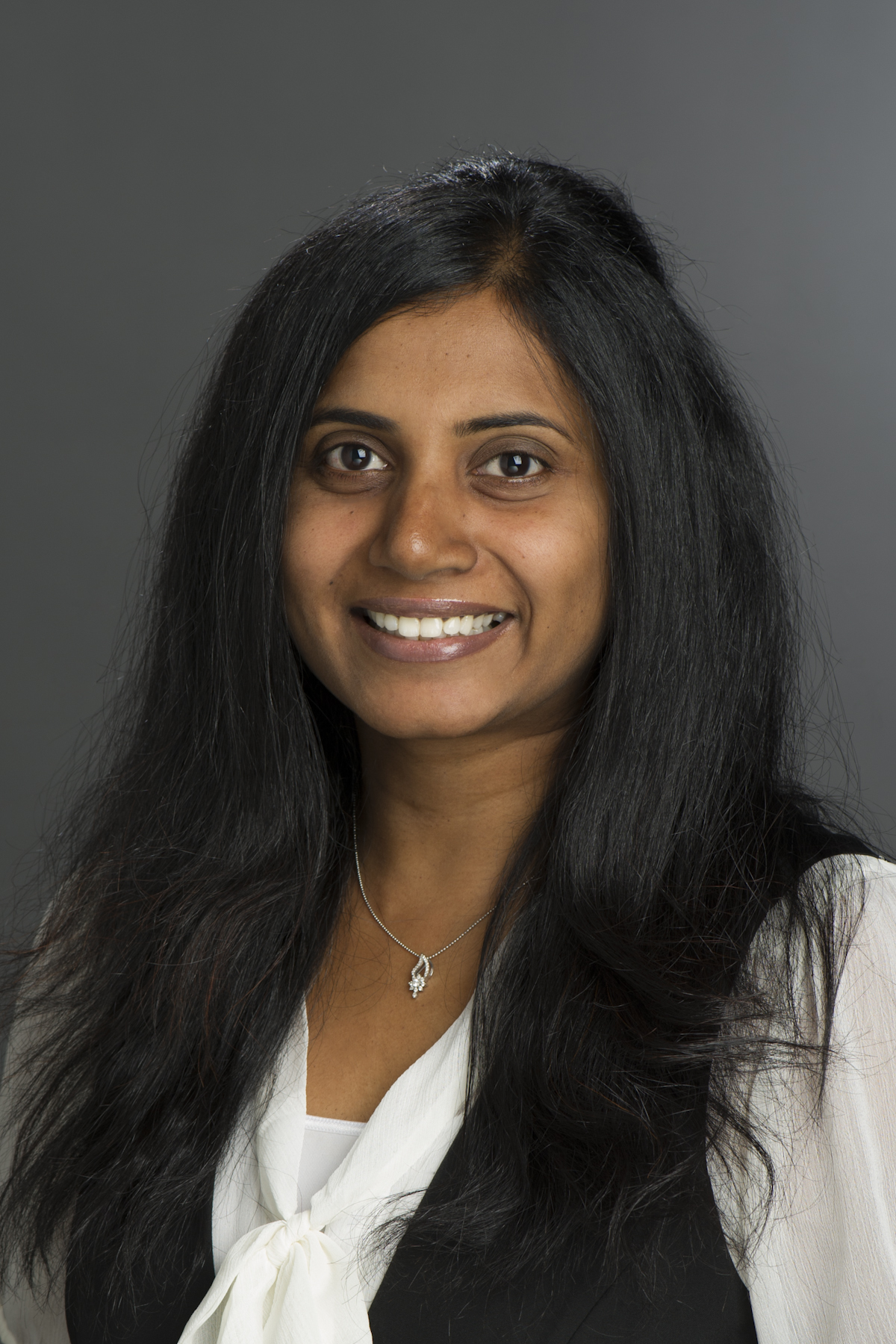}}]{Dr. Dhireesha Kudithipudi}[M'06, SM'16] is a professor and founding Director of the AI Consortium at the University of Texas, San Antonio and Robert F McDermott Chair in Engineering. Her research interests are in neuromorphic AI, low power machine intelligence, brain-inspired accelerators, and use-inspired research. Her team has developed comprehensive neocortex and cerebellum based architectures with nonvolatile memory, hybrid plasticity models, and ultra-low precision architectures. She is passionate about transdisciplinary and inclusive research training in AI fields. She is the recipient of the Clare Booth Luce Scholarship in STEM for women in highered (2018) and the 2018 Technology Women of the Year in Rochester.
\end{IEEEbiography}

\end{document}

%% file: Config.tex
\usepackage{booktabs} 
\usepackage{subfigure}
\usepackage{blindtext}
\usepackage{graphicx}
\usepackage{csquotes}
\usepackage{caption}
\usepackage{cite}
\usepackage{amssymb}
\usepackage{amsmath}
\usepackage{microtype}
\usepackage{multicol}
\usepackage{upgreek}
\usepackage{booktabs,tabularx,ragged2e,url}
\usepackage[english]{babel}
\usepackage{fancyhdr}
\usepackage[noend]{algpseudocode}
\usepackage{float}
\usepackage{blindtext}
 \usepackage{siunitx}

\usepackage{academicons}
\usepackage{xcolor}
\newcommand{\orcid}[1]{\href{https://orcid.org/#1}{\textcolor[HTML]{A6CE39}{\aiOrcid}}}

\usepackage{nomencl}
\makenomenclature

\usepackage[ruled,linesnumbered,commentsnumbered]{algorithm2e}

\SetAlFnt{\small}
\SetAlCapFnt{\small}
\SetAlCapNameFnt{\small}
\SetAlCapHSkip{0pt}
\IncMargin{-\parindent}

\usepackage[hang,flushmargin]{footmisc}

\usepackage{xcolor,colortbl}
\definecolor{Gray}{gray}{0.90}
\newcolumntype{a}{>{\columncolor{Gray}}c}

\usepackage{xcolor,soul}
\definecolor{light-gray}{gray}{0.95}
\sethlcolor{light-gray}

\graphicspath{{Figures/}}
\newcommand{\fig}[1]{Fig.~\ref{#1}}
\newcommand{\tb}[1]{Table~\ref{#1}}
\newcommand{\eq}[1]{(\ref{#1})}

\usepackage{enumitem,booktabs}
\usepackage[referable]{threeparttablex}
\renewlist{tablenotes}{enumerate}{1}
\makeatletter
\setlist[tablenotes]{label=\tnote{\alph*},ref=\alph*,itemsep=\z@,topsep=\z@skip,partopsep=\z@skip,parsep=\z@,itemindent=\z@,labelsep=.2em,leftmargin=*,align=left,before={\footnotesize}}
\makeatother

\usepackage[font={small}]{caption}

\DeclareCaptionLabelSeparator{periodspace}{.\quad}
\usepackage{enumitem,amssymb}
\newlist{todolist}{itemize}{2}
\setlist[todolist]{label=$\square$}
\usepackage{pifont}